\documentclass[sigconf]{acmart}

\AtBeginDocument{%
  \providecommand\BibTeX{{%
    \normalfont B\kern-0.5em{\scshape i\kern-0.25em b}\kern-0.8em\TeX}}}

\setcopyright{acmcopyright}
\copyrightyear{2018}
\acmYear{2018}
\acmDOI{10.1145/1122445.1122456}

\acmConference[Woodstock '18]{Woodstock '18: ACM Symposium on Neural
  Gaze Detection}{June 03--05, 2018}{Woodstock, NY}
\acmBooktitle{Woodstock '18: ACM Symposium on Neural Gaze Detection,
  June 03--05, 2018, Woodstock, NY}
\acmPrice{15.00}
\acmISBN{978-1-4503-XXXX-X/18/06}

\renewcommand\footnotetextcopyrightpermission[1]{} 
\usepackage{layouts}
\usepackage{titlesec}
\titlespacing*{\section}{0pt}{0.7\baselineskip}{0.25\baselineskip}
\titlespacing*{\subsection}{0pt}{0.7\baselineskip}{0.18\baselineskip}
\titlespacing*{\subsubsection}{0pt}{0.5\baselineskip}{0.15\baselineskip}

\usepackage{fancyhdr}
\usepackage[normalem]{ulem}

\usepackage{algorithmic}
\usepackage{graphicx}
\usepackage{textcomp}
\usepackage{xcolor}
\usepackage{todonotes}

\usepackage{xurl}
\usepackage{etoolbox}

\usepackage{titlesec}
\usepackage{lipsum}
\usepackage{layouts}

\usepackage[utf8]{inputenc}

\usepackage{float}
\usepackage{booktabs} 
\usepackage{multirow}

\usepackage{caption}
\usepackage{subcaption}

\usepackage{makecell}

\usepackage{tabularx,ragged2e,booktabs,caption}

\usepackage{mathtools}

\usepackage{balance}
\usepackage{listings}
\usepackage{color}
\definecolor{dkgreen}{rgb}{0,0.6,0}
\definecolor{gray}{rgb}{0.5,0.5,0.5}
\definecolor{mauve}{rgb}{0.58,0,0.82}
\definecolor{bg}{rgb}{0.9,0.9,0.9}
\definecolor{amber}{rgb}{1.0, 0.75, 0.0}

\PassOptionsToPackage{hyphens}{url}\usepackage{hyperref}

\newcommand{\speedupcpu}{18.2}
\newcommand{\speedupgpu}{26.9}
\newcommand{\framework}{SPA-GCN}
\usepackage{pifont}
\newcommand{\xmark}{\ding{55}}

\usepackage{enumitem}
\setlist[itemize,1]{itemsep=0.5pt,partopsep=0pt,parsep=\parskip, topsep=2pt, leftmargin=10pt,}
\setlist[enumerate,1]{itemsep=0.5pt,partopsep=0pt,parsep=\parskip, topsep=2pt, leftmargin=10pt,}

\begin{document}
\title{SPA-GCN: Efficient and Flexible GCN Accelerator with an Application for Graph Similarity Computation}

\author{Atefeh Sohrabizadeh, 
        Yuze Chi, and Jason Cong}
\affiliation{%
  \institution{Computer Science Department, University of California, Los Angeles, USA}
}
\email{{atefehsz, chiyuze, cong}@cs.ucla.edu}
\pagenumbering{arabic}

\thispagestyle{firstpage}
\pagestyle{plain}


\begin{abstract}
    While there have been many studies on hardware acceleration for deep learning on images, there has been a rather limited focus on accelerating deep learning applications involving graphs. The unique characteristics of graphs, such as the irregular memory access and dynamic parallelism, impose several challenges when the algorithm is mapped to a CPU or GPU. 
    To address these challenges while exploiting all the available sparsity, we propose a flexible architecture called \mbox{\framework} for accelerating Graph Convolutional Networks (GCN), the core computation unit in deep learning algorithms on graphs. The architecture is specialized for dealing with many small graphs since the graph size has a significant impact on design considerations.
    In this context, we use SimGNN, a neural-network-based graph matching algorithm, as a case study to demonstrate the effectiveness of our architecture.
    The experimental results demonstrate that \mbox{\framework} can deliver a high speedup compared to a multi-core CPU implementation and a GPU implementation, showing the efficiency of our design.
\end{abstract}

\maketitle


\section{Introduction} \label{sec:intro}
Graphs are the core data structure used in datacenters and have a wide application in different domains such as recommender systems, social networks, and the World Wide Web.
Although they are widely used, they are mainly unstructured and have a high dimensionality, making them computationally expensive to process.
In fact, many graph algorithms, such as Graph Edit Distance~\cite{ged} and Maximum Common Subgraphs~\cite{mcs} are known to be NP-complete~\cite{zeng2009comparing, kann1992approximability}.
This problem has motivated researchers to apply deep learning on graphs with the goal of extracting structured, low-dimensional features from it (e.g., ~\cite{gcn, ying2018graph}). More specifically, the purpose of this line of research is to assign a feature vector to each node of the graph that can show the role of the node in the graph.
Such feature vectors are called the \textit{node embeddings}.

In this context, Graph Convolutional Networks (GCN)~\cite{gcn} are widely used to extract the node embeddings in a graph. GCNs follow the same behavior as Convolutional Neural Networks (CNN) in learning. They consist of multiple layers in which the features of the nodes are propagated within them until rich information of the input graph is derived. Like in a CNN, in each layer, the GCN updates the node features by gathering the neighbors' features and passing their summation through a filter. GCNs have shown to be successful in many domains including traffic prediction~\cite{zhao2019t}, facilitating web-scale recommender systems~\cite{ying2018graph}, molecular footprint calculation~\cite{duvenaud2015convolutional}, logic optimization for EDA tools~\cite{haaswijk2018deep}, etc. 

While some graph data tend to scale rapidly, there are also many graph data that are naturally limited in size, for example, chemical compounds and molecules~\cite{aids, zinc, berman2000protein, bolton2008pubchem, chen2019alchemy} that have a wide application in different domains including drug development, quantum mechanics, physical chemistry, biophysics, etc~\cite{wu2018moleculenet, chen2019alchemy}; the LINUX dataset containing operating system program dependence graph~\cite{wang2012efficient}; the GREC database consisting of graphs representing symbols from architectural and electronic drawings~\cite{grec}; the Fingerprint database~\cite{fingerprint}, etc~\cite{wu2018moleculenet, riesen2008iam}. The average number of nodes for the graphs of these databases ranges from 5 to 50.

Because of the vast application of small graphs, numerous algorithms have been proposed to obtain their information by extracting their features or learning how similar two graphs are~\cite{ishida2021graph, ma2020multi, chen2019efficient, kim2019inves, qin2020ghashing, gmn, bai2019simgnn}, especially using GCNs~\cite{ishida2021graph, gmn, ma2020multi,qin2020ghashing, bai2019simgnn}. In particular, SimGNN~\cite{bai2019simgnn} proposed a GCN-based approach to learn a similarity score for such graphs. It demonstrates that a GCN-based approach is able to approximate the GED with high accuracy; hence, it expedites the graph similarity computation significantly for many applications. SimGNN targets searching/matching graphs from real-world graph databases, such as AIDS~\cite{aids}, LINUX~\cite{wang2012efficient}, and IMDB~\cite{yanardag2015deep}, which consists of antivirus chemical compounds, program dependency graphs, and actor/actress ego-networks, respectively. The generated target graphs are relatively small, with 10 nodes on average, but the database contains millions of graph pairs, creating a large number of graph matching queries. Although the CPU implementation can finish each SimGNN query for such graphs in milliseconds (5.8ms to be exact), processing millions of queries can take several hours, which is not acceptable and requires customized acceleration. Such a workload of graph searching/mining is increasing in importance. For example, searching for antivirus chemical compounds is an important step in drug repurposing for COVID-19.  

\begin{table*}[!tbh]
\footnotesize
\centering
\caption{Our approach compared to state-of-the-art GCN accelerators.}
\label{tbl:comparison}
\begin{tabular}{ccccccccc}
\hline
\multirow{2}{*}{\textbf{Work}} & \multirow{2}{*}{\makecell{\textbf{Graph}\\ \textbf{Size}}}   & \multirow{2}{*}{\makecell{\textbf{Layer}\\ \textbf{Customization}}} & \multirow{2}{*}{\makecell{\textbf{Sparse Engine for} \\ \textbf{Feature}\\ \textbf{ Transformation}}}    & \multirow{2}{*}{\makecell{\textbf{Read Each} \\ \textbf{Element} \\ \textbf{Only Once}}}  & \multicolumn{4}{c}{\textbf{Parallelization}}\\\cline{6-9}
    &&&  &  &\textbf{Inter-layer} & \makecell{\textbf{Feature-level} \\  \textbf{(Sparse Part)}} & \makecell{\textbf{Node-level} \\  \textbf{(Sparse Part)}} & \textbf{Batch}\\\\ \hline\hline
HyGCN~\cite{yan2020hygcn} & Large & \xmark & \xmark & \xmark & \xmark & \checkmark & \checkmark & \xmark\\
GraphACT~\cite{zeng2020graphact} & Small & \xmark & \xmark & \xmark & \xmark &  \checkmark& \xmark &\xmark\\
ASAP'20~\cite{zhang2020accelerating} & Large & \xmark  & \xmark & \xmark & \xmark & \checkmark & Limited &\xmark \\
AWB-GCN~\cite{geng2020awb} & Large & \checkmark & \checkmark & \xmark & \checkmark & \xmark & \checkmark  & \xmark\\
\hline
\textbf{Ours (\mbox{\framework})} &  Small & \checkmark & \checkmark & \checkmark & \checkmark & \checkmark & \checkmark & \checkmark \\
\hline
\end{tabular}
\end{table*}


Despite the popularity and effectiveness of \textit{graph neural network} (GNN) approaches, there has been limited research on developing an accelerator for them. Besides, the differences between an image and a graph structure, as explained below, make the countless CNN accelerators proposed in the literature (e.g., ~\cite{zhang2018dnnbuilder, wei2017automated, wei2018tgpa, zeng2018framework, zhang2018caffeine, shen2017escher, flexcnn, nakahara2020high, mo2020tfe, sharma2018bit, chen2016eyeriss, judd2016stripes, shao2019simba, song2017pipelayer}) incompatible. Furthermore, the characteristics of the graphs impose several challenges when the algorithms are executed using CPUs or GPUs. More specifically, the unique features of GNN impose the following challenges in designing an accelerator:
\begin{itemize}
    \item \textbf{Irregular memory access and low data reuse:} As opposed to images in which the neighbors of a pixel are stored either in contiguous locations or have a fixed distance to one another, the neighbors of a node in a graph may be stored in any location in memory. This will result in many irregular memory accesses to all levels of the memory hierarchy. To make matters worse, GNNs have much lower data reuse compared to CNNs. These characteristics make the application memory-bounded. Compared to the traditional graph algorithms such as breadth first search (BFS), single-source shortest path (SSSP), PageRank (PR), etc., the nodes have long feature vectors instead of a single scalar value. As a result, although they both have irregular memory access, the access pattern is different. Also, this characteristic introduces new kinds of parallelism and data reuse. Now that each node deals with a long vector (often length of 256 or higher) rather than a scalar, we can exploit intra-node parallelism. Furthermore, all the vectors associated with different nodes share a weight matrix which brings in data reuse opportunities. These differences make most graph-based accelerators (e.g., ~\cite{dai2017foregraph, dai2016fpgp, zhou2016high, chi2016nxgraph, zhou2015accelerating, Graphicionado, wang2019processor, isca17-aggressive-pipelining, micro18-hast, Basak2019, isca20-graphabcd, Dadu2019, wang2016gunrock, hlrc}) ill-suited for GNNs.
    \item \textbf{Computation pattern disparity:} Different steps of the GCN algorithm deal with different sparsity rates as will be discussed in Section~\ref{sec:gcn}.
    In addition, a GNN may utilize other types of computation patterns, such as neural tensor network computation in SimGNN (see Section~\ref{sec:simgnn} for details) to make an end-to-end application. Such computation pattern variations call for a customized processing unit for each of these steps.
    \item \textbf{Dynamic workload and parallelism:} Apart from the random memory location of the neighbors of a node, the \textit{number} of neighbors also varies across different nodes. This will result in load-imbalance between the graph's nodes that can degrade the performance significantly.
\end{itemize}

In addition to the challenges mentioned above, dealing with small graphs (which is particularly challenging for GPUs as shown in Section~\ref{sec:comparisons}) requires special design considerations as we will explain in Section~\ref{sec:arc_gcn}.
To solve these challenges, in this paper, we present \textit{\mbox{\framework}} as an efficient and flexible \underline{\textit{\textbf{GCN}}} \underline{\textit{\textbf{a}}}ccelerator for small graphs that exploits all the available \underline{\textit{\textbf{sp}}}arsity. 
Then, we apply it to accelerate the entire processing pipeline of SimGNN as an end-to-end application. Since we are facing a \textit{memory-bounded} application, our goal is to have the least number of global memory transactions and to read each element only once. Because of the \textit{computation pattern disparity}, we analyze the requirements for each step of the computation pipeline and, accordingly, develop a dedicated architecture for each of them. To deal with \textit{the irregular memory access}, we utilize a scratchpad memory to store the matrices that need random access. This is doable as we are targeting small graphs. We further propose an efficient workload distribution mechanism to alleviate the \textit{load imbalance} problem. 

Concisely, we fuse all the stages together and employ a very deep pipeline with four different levels of nested parallelization and customize the computation units for each of the stages based on its workload as listed in Table~\ref{tbl:comparison}. These design decisions distinguish \mbox{\framework} from recently proposed GCN accelerators~\cite{zeng2020graphact, yan2020hygcn, geng2020awb, zhang2020accelerating} as summarized in Table~\ref{tbl:comparison}. Section~\ref{sec:rel-work} compares our approach to prior works in more details. 
While we use SimGNN for illustrating our approach, the same optimizations can be applied to other GCN-based networks dealing with small graphs such as~\cite{qin2020ghashing, ishida2021graph, graphsim} as well. 
We implement \mbox{\framework} on three different FPGAs showing the flexibility and adaptivity of \mbox{\framework} to different platforms with different global memory bandwidth.
Experimental results suggest that \mbox{\framework} mapped to an HBM FPGA outperforms a multi-core CPU and a GPU implementation by {\speedupcpu}$\times$ and {\speedupgpu}$\times$, respectively.

In summary, the key contributions of this paper are:
\begin{itemize}
    \item We design and develop \mbox{\framework}, a flexible architecture for accelerating GCN specialized for processing many small graphs.
    \item We adopt \mbox{\framework} to accelerate SimGNN as an end-to-end application, resulting in an efficient architecture with a very deep pipeline and four levels of parallelization. To our knowledge, this is the first hardware accelerator for GCN-based graph matching.
    \item We demonstrate the flexibility of our architecture by mapping and customizing it to three different FPGAs with different capacities and memory systems.
    \item Experimental results suggest that our accelerator can outperform multi-core CPU by {\speedupcpu}$\times$ and GPU by {\speedupgpu}$\times$, demonstrating the efficiency of our design.
\end{itemize}

\section{Background} \label{sec:bg}
\subsection{Graph Convolutional Network (GCN)} \label{sec:gcn}
Inspired by the success of Convolutional Neural Networks (CNN) on images, GCN~\cite{gcn} was developed to apply a series of convolutional layers on graphs. Layer \textit{l} of a GCN takes an undirected graph $G(V, E, H^l)$ as the input, where \textit{V} and \textit{E} denote the nodes and edges of the graph, respectively. $H^l \in \mathbb{R}^{|V|\times f_{l}}$ is the matrix of the \textit{input node embeddings} for this layer, with each row containing the embedding (also called node features in the literature) of one of the nodes where $f_l$ indicates the number of features of each node at layer $l$. The core computation of a GCN layer to produce the \textit{output node embeddings}, $H^{l+1} \in \mathbb{R}^{|V|\times f_{l+1}}$ is as follows:
\begin{align} \label{eq:gcn}
    \begin{split}
        H^{l+1} = \sigma_{activation} \left(A' \cdot H^l \cdot W^l \right)
    \end{split}
\end{align}
where $\sigma_{activation}(\cdot)$ is an activation function which typically is a ReLU and $W^l$ is a layer-specific \textit{trainable weight matrix}. $A'$ is the \textit{normalized adjacency matrix with added self-connections} and can be calculated as follows:
\begin{align} \label{eq:adjacency-mat}
    \begin{split}
        \Tilde{A} = A &+ I_{N} , \quad
        \Tilde{D}_{ii} = \sum_j \Tilde{A}_{ij}, \quad 
        A' = \Tilde{D}^{-\frac{1}{2}} \cdot \Tilde{A} \cdot \Tilde{D}^{-\frac{1}{2}}
    \end{split}
\end{align}
here, $A$ and $I_N$ are the \textit{adjacency} and the \textit{identity matrix}, respectively. $\Tilde{D}$ is a \textit{diagonal matrix} where $\Tilde{D}_{ii}$ is the degree of node \textit{i} after the self-connection edges are added.

\begin{figure}[!htb]
	\centering
	\includegraphics[width=0.6\columnwidth]{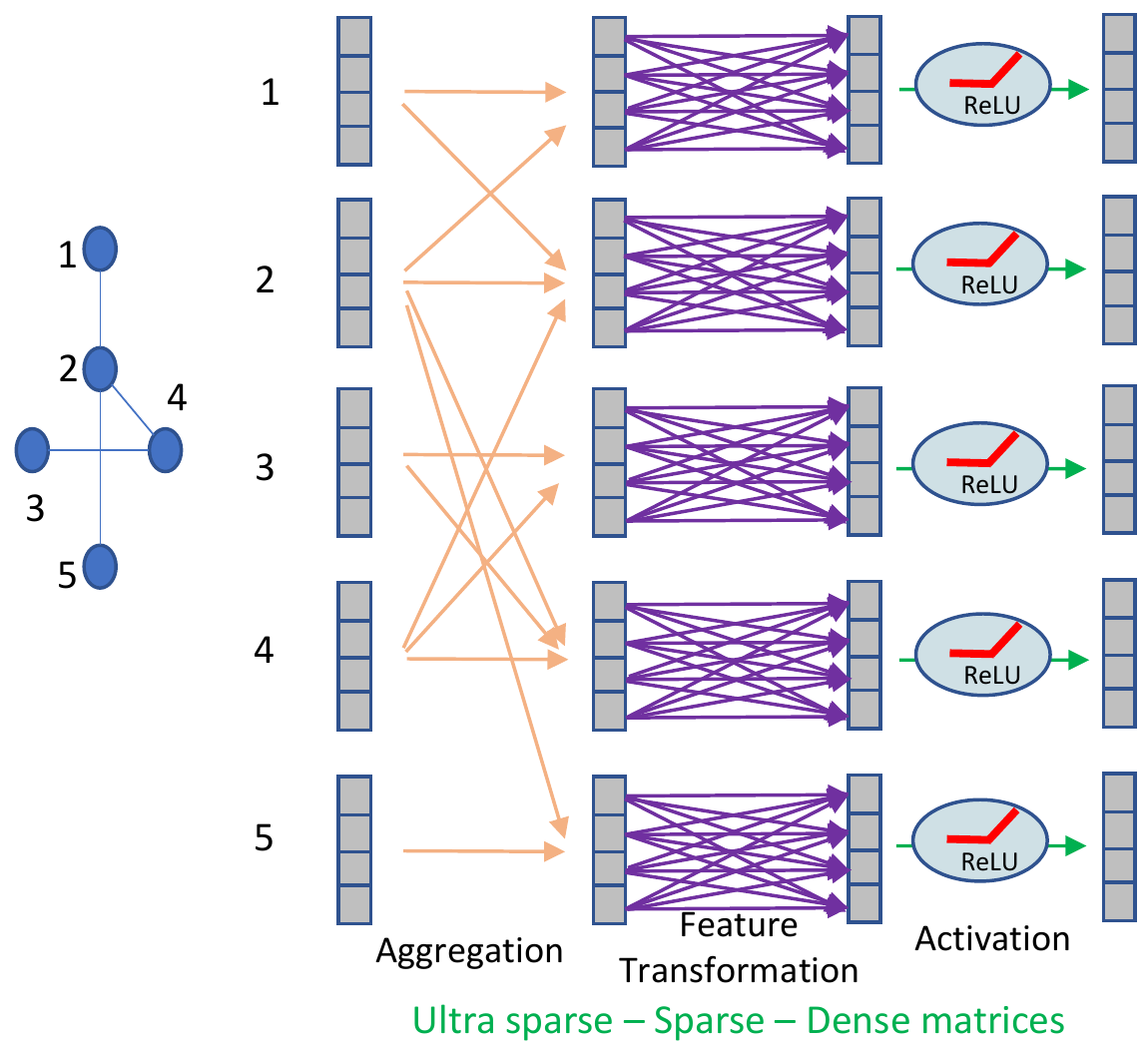} 
	\caption{Overview of the core computation in a GCN Layer}
	\label{fig:gcn}
\end{figure}

Fig.~\ref{fig:gcn} depicts the computation in Eq.~\ref{eq:gcn} on a simple toy graph assuming that $A'$ is given, the activation function is ReLU, and each node embedding is a vector of size 4. The first step in the computation gathers the information from the neighbors of each of the nodes which is denoted in Eq.~\ref{eq:gcn} as $A' \cdot H^l$. Since $A'$ is a normalized matrix, the computation here is a weighted aggregation. After the Aggregation step, the node embeddings are transformed by applying a pre-trained set of weights. Note that the output embeddings may have a different vector size. In the end, the embeddings are passed through a ReLU layer to introduce non-linearity to the model. The time complexity for each GCN layer is $O(|E|f_{in}f_{out})$, where $|E|$ denotes the number of edges including the self-connection ones, since $A' \cdot H^l$ can be implemented as a sparse-dense matrix multiplication~\cite{gcn}. As we will explain in Section~\ref{sec:gcn-sparse}, because of the ReLU unit at the end of each GCN layer, the node embeddings will also be a sparse matrix. However, the rate of sparsity is different as the adjacency matrix is, in fact, an \textit{ultra} sparse matrix~\cite{geng2020awb}.

\section{\mbox{\framework} Architecture} \label{sec:arc_gcn}
The first step in designing an architecture for GCN is to determine the order of the matrix multiplications. More specifically, we can either compute Eq.~\ref{eq:gcn} as $(A'\times H^l)\times W^l$ or $A'\times (H^l \times W^l)$. We have chosen the latter since it results in a fewer number of operations. Intuitively, this is because both matrices $A'$ and $H^l$ are sparse, but their multiplication creates a dense matrix. As a result, in the former, we end up doing a dense-dense multiplication for the second multiplication. However, if we go with the latter, both multiplications are sparse-dense that as shown in AWB-GCN~\cite{geng2020awb}, reduces the number of operations. 
Figure~\ref{fig:high-level-gcn} illustrates the high-level view of GCN architecture in \mbox{\framework}.
In this section, we employ a bottom-up approach to highlight the optimization opportunities when GCN is applied to small graphs and how we used them to build the GCN accelerator as demonstrated in Figure~\ref{fig:high-level-gcn}.

\subsection{\mbox{\framework} Design Principles}
As our application is memory-bounded and we are targeting small graphs, \mbox{\framework} is designed:
\begin{itemize}
    \item To reduce the number of times we access the global memory to the least amount possible. In our final architecture, each input element is read only once and there is no need to store any of the intermediate results in the global memory.
    \item To exploit all the available sparsity.
    \item To employ a deep pipeline with varying levels and degrees of parallelization for matching the workload of different stages and maximizing the overall performance.
    \item To customize the architecture to efficiently handle small graphs.
\end{itemize}

\subsection{Baseline Architecture} \label{sec:gcn_baseline}
In this section, we describe the basic optimizations that can be applied for processing GCNs. 
We explain the optimized way of scheduling the operations of a GCN when targeting small graphs and justify the use of an outer-product-based multiplication~\cite{van1997summa, buluc2008representation} for it. Furthermore, we demonstrate how we support sparse computation in the Aggregation step. Although these optimizations are necessary, they are not enough when dealing with many small graphs. We cover the rest of the optimizations that can be applied here in the subsequent sections.

\subsubsection{Feature Transformation:} \label{sec:feature_transform}

In this step, one has to multiply matrices $H^l \in \mathbb{R}^{|V| \times f_{in}}$ and $W^l \in \mathbb{R}^{f_{in} \times f_{out}}$, where $f_{in}$ and $f_{out}$ denote the number of input and output features, respectively. Since the matrix multiplication has $|V| \times f_{in} \times f_{out}$ operations, the minimum latency for doing this computation is: $\frac{|V| \times f_{in} \times f_{out}}{\#MAC_{used}}$ where $\#MAC_{used}$ is the number of used \textit{multiply and accumulate} (MAC) units. To achieve this latency, we should be able to schedule a new set of operations in each clock cycle. However, adopting an inner-product-based matrix multiplication results in updating the same output feature in the consecutive iterations which introduces read-after-write a (RAW) dependency between them. As a result, our pipeline cannot achieve an initiation interval (II) of one which degrades the efficiency of our accelerator.

\textit{\textbf{Optimized Scheduling}: Read the Weight Matrix Row-wise; Stream the Embeddings Matrix Column-wise.} We can solve the aforementioned problem by changing the order of computations. To break the dependency, we should update different output locations by taking an element from one of the input matrices (read as a stream) and broadcasting that to parallel MAC units while each MAC unit reads different elements from the other matrix.
We choose to read $H^l$ as a stream and prefetch and cache $W^l$ since it will be reused by $H^l$. With this scheduling we can support sparsity for this step more efficiently as discussed in Section~\ref{sec:gcn-sparse}.

\begin{figure*}[!phtb]
	\centering
	\includegraphics[width=0.75\linewidth]{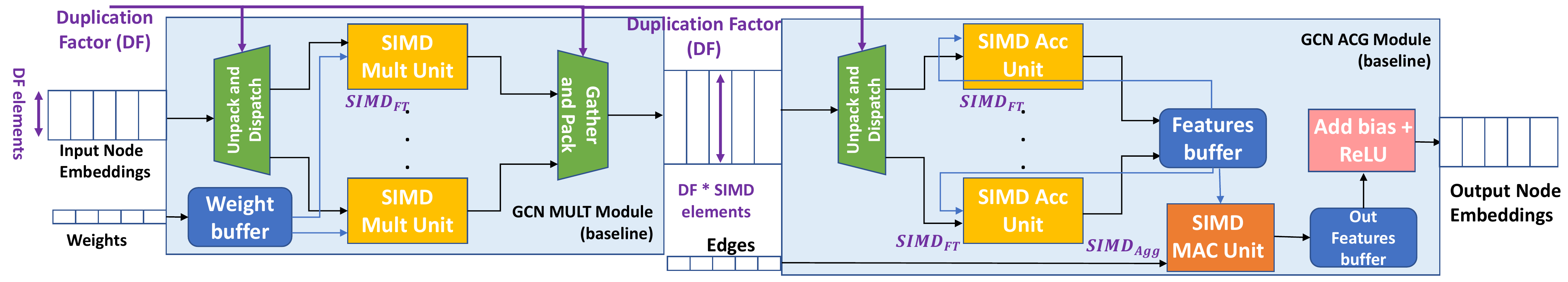} 
	\caption{\mbox{\framework} baseline architecture: intra-layer pipelining between MULT module (multiplication unit for Feature Transformation step) and ACG Module (accumulation unit for Feature Transformation step + Aggregation step)}
	\label{fig:baseline-arc}
\end{figure*}
\begin{figure}[!htb]
	\centering
	\includegraphics[width=\columnwidth]{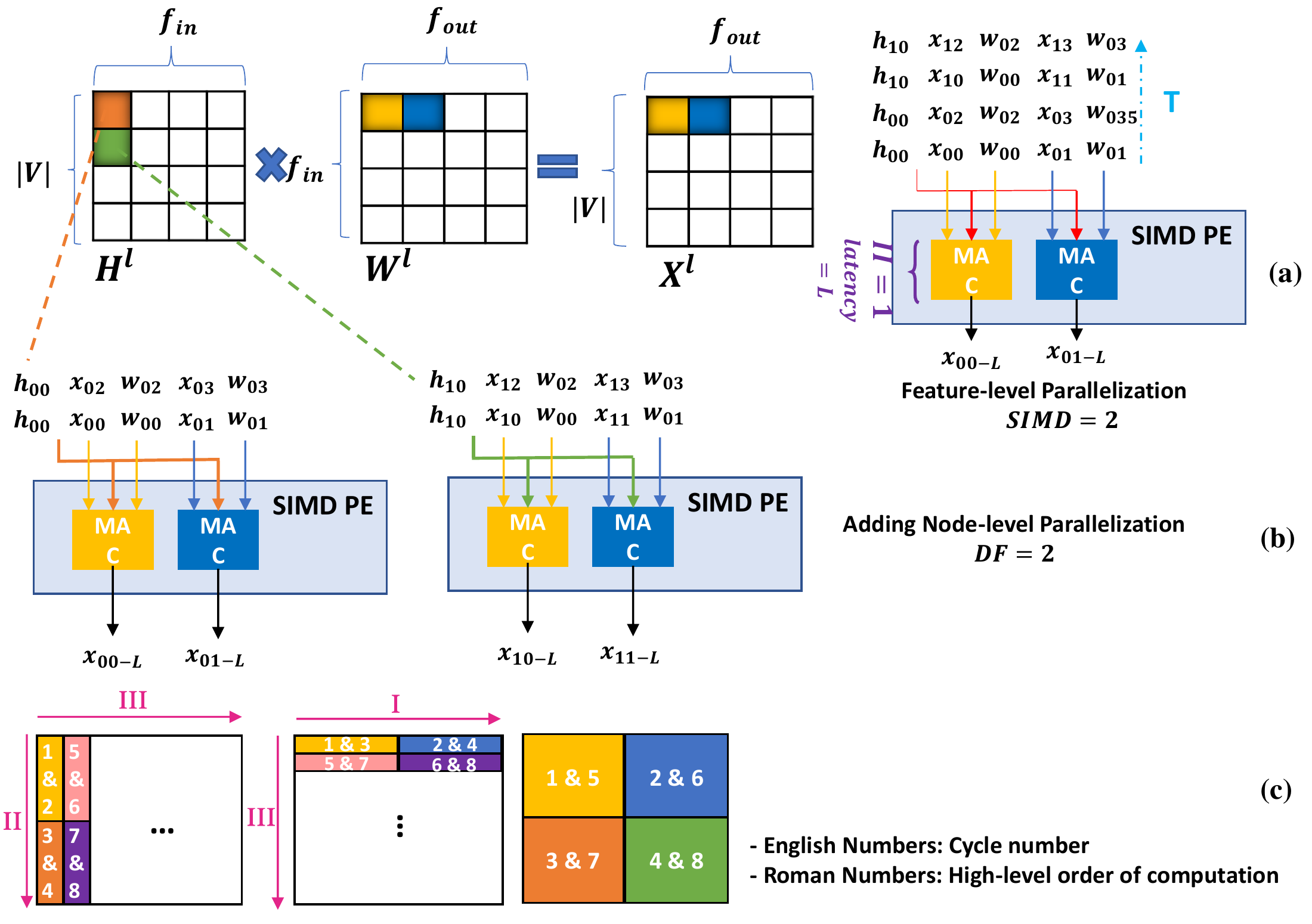} 
	\caption{Scheduling order for Feature Transformation: (a) Feature-level parallelization. (b) Node-level parallelization. (c) The overall computation order.}
	\label{fig:mm-sched}
\end{figure}

Fig.~\ref{fig:mm-sched}~(a) depicts the modified scheduling for cycle \textit{l}. As the figure suggests, we divide the workload within a PE by parallelizing \textit{SIMD} operations across the output features. To read each element only once, for each fetched element of $H^l$, (i.e., $h_{00}$), we schedule all the operations it is involved with, before its eviction. 
In addition to reusing the $h_{00}$, this scheduling increases the number of cycles before a RAW dependency happens. This is because by reading each element of the input node embedding, we can ensure that we will be updating different output locations in the next $\frac{f_{out}}{SIMD}$ cycles. 
$x_{ij-L}$ elements in the figure represent the result of the MAC operations that were scheduled \textit{L} cycles (\textit{L} being the latency of a MAC unit) before the current inputs: $x_{ij}$ elements. As long as $\forall l \in [1\,..\,L-1]\; x_{ij-l} \neq x_{ij}$, there is no dependency; hence, we can schedule a new operation every cycle and achieve II=1. To make sure that II=1 is possible,
we read matrix $H^l$ in column order. Note that if we read it rather in row order, we update the same location every $\frac{f_{out}}{SIMD}$ iterations instead of every $|V| \times \frac{f_{out}}{SIMD}$ iterations. 

We add a second level of parallelization by duplicating the SIMD PE by a \textit{duplication factor (DF)} which parallelizes the node dimension. As a result, we can make use of memory coalescing by packing and reading DF elements in each cycle. 
Fig.~\ref{fig:mm-sched}(c) illustrates the final execution order of this step. The arrows denote the high-level ordering of traversing different dimensions and the numbers show the elements that are accessed at their respective cycles. It is important to traverse the input feature dimension ($f_{in}$) last (\textit{arrow} $III$) since it is the dimension causing the dependencies. In the baseline architecture, to ensure that the dependency distance is larger than \textit{L} (to get II=1), we pad the matrix $H^l$ with zeros by adding rows until $\frac{(|V| + padding)}{DF} \times \frac{f_{out}}{SIMD} \ge L$. In Section~\ref{sec:gcn-sparse}, however, we insert bubbles in case of a dependency since the location of non-zero elements is dynamic.

\vspace{0.1in}
\subsubsection{Aggregation:} \label{sec:agg}
In this step, we must multiply matrices $A' \in \mathbb{R}^{|V| \times |V|}$ and $X^l \in \mathbb{R}^{|V| \times f_{out}}$ where $X^l$ is the result of the Feature Transformation step and $A'$ denotes the normalized adjacency matrix with added self-connections. Due to the highly irregular access to the matrix $X^l$ to aggregate features of the neighbors, we allocate a scratchpad memory to it. 
Matrix $A'$, is often ultra sparse~\cite{geng2020awb}. To reduce the number of both transferred elements and operations, we prune this matrix and only pass its non-zero elements, which represent edges, to FPGA. For most graph databases, like the ones we target, this step is not needed since the graphs are already stored as a data structure that contains a list of the vertices and edges. Instead of dedicating an on-chip memory for storing the edges, we read them as a stream and update all the features of the destination node, before retiring the edge. 

We further pre-process the adjacency matrix to not only pre-compute $A'$, but also re-organize the edges to prevent having RAW dependencies. More specifically, before sending the edges to the FPGA, they are re-arranged offline so that the ones with the same destination node are at least \textit{L} locations apart to make sure there is not more than one update to the same node within the window of \textit{L} cycles. Without this re-ordering, we either have to increase the II for the whole operation or add a control unit to insert bubbles in the pipeline of this computation engine in case there is a RAW dependency to ensure correctness. We chose to pre-process the edges since as we are dealing with small graphs, the time for this pre-processing is negligible. We only make use of feature-level parallelism to distribute the workload in this step. Edge-level parallelism can result in bank conflicts since they update random nodes.

\subsubsection{Intra-layer Pipelining:} \label{sec:gcn-intra-p}
In the last two sections, we explained the basic optimizations that should be applied to the major computation steps of GCN, Feature Transformation and Aggregation. The last step (ReLU) has a very low area and latency overhead. We only utilize a $max(0, \cdot)$ unit at the end of the aggregation module for it. To further boost the performance, we add intra-layer pipelining by exploiting a dataflow architecture for connecting the modules (see Fig.~\ref{fig:baseline-arc}). As a result, the overall latency will be close to the latency of the slowest module (here, the ACG module). In addition, we can avoid off-chip memory accesses in between these modules.

The computation in the Feature Transformation step can be divided into two separate tasks. Task 1, the MULT unit, is responsible for doing all the multiplications and Task 2, the ACC unit, does the additions. We pipeline these tasks by implementing them as separate modules that are connected by FIFOs. To avoid deadlocks, the throughput (i.e. II) of these units should match. The MULT module, as depicted in Fig.~\ref{fig:baseline-arc}, has a local buffer to store the weights and streams the elements of $H^l$ from input FIFO. Each entry of this FIFO is a concatenation of DF elements (see Section~\ref{sec:feature_transform}). 
Both the SIMD factor and DF can be customized based on the target network configuration. Once the multiplication results are ready, they are packed and sent to the ACG module.

The ACC unit of the Feature Transformation step and the Aggregation step share the matrix $X^l$; 
thus, to save memory resources, we merge their computation into one module as shown in \textit{ACG} module in Fig.~\ref{fig:baseline-arc}. After fetching the output of the \textit{MULT} module, the \textit{ACG} module unpacks the data based on the same DF, and dispatches SIMD elements to each \textit{SIMD Acc Unit} with the same SIMD factor. Once the additions are done, it will store the partial results to the local buffer \textit{features buffer}. After all updates are committed to the \textit{features buffer}, the matrix $X^l$ is computed and the Aggregation step can start. 
The SIMD factor of this step is higher than the one in Feature Transformation step since we only exploit feature-level parallelization here. After this step is finished, the elements of the \textit{out features buffer} are added with a bias, passed through a ReLU unit, and stored into off-chip memory. Note that in the baseline architecture, we reuse the same modules for all the GCN layers.

\begin{figure*}[!phtb]
	\centering
	\includegraphics[width=\linewidth]{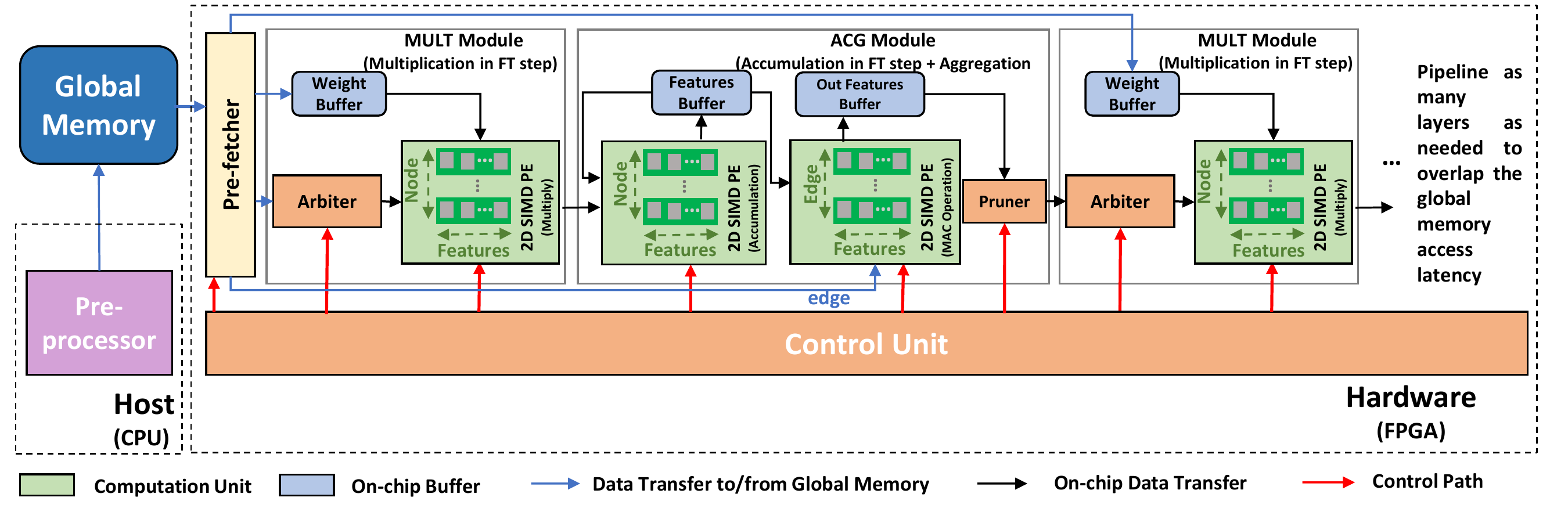} 
	\caption{High-level overview of GCN accelerator architecture in \mbox{\framework}}
	\label{fig:high-level-gcn}
\end{figure*}


\subsection{Adding Inter-layer Pipelining} \label{sec:gcn-TLP}
As it is commonly practiced (\cite{yan2020hygcn, zeng2020graphact, zhang2020accelerating, liang2020engn}), in the baseline architecture, we only exploit intra-layer pipelining and reuse the modules for all the GCN layers. However, this is not sufficient when we are dealing with small graphs. The off-chip communication is a serious burden for this application since it deals with small-sized inputs. To alleviate this problem, we intend to reduce the number of accesses to the off-chip memory as much as possible. The baseline architecture is inefficient with this regard since, at the end of each layer, the output should be stored to the off-chip memory and read back again for the next layer. To avoid these redundant accesses, we propose to extend the dataflow architecture described in Section~\ref{sec:gcn-intra-p} to all layers of GCN. To realize this, we instantiate new modules for each layer and connect them with FIFOs as depicted in Fig.~\ref{fig:high-level-gcn}.

The various (LUT, BRAM, and URAM) and large on-chip memory resources of FPGA and the rather small matrices used for our target application enable us to dedicate separate modules for each of the layers.
Fusing the computation for all the layers by enabling dataflow architecture has several benefits such as: 1) we can avoid writing the intermediate results to the global memory by forwarding them to the next layer through FIFOs. 2) The operations will be dynamically scheduled since each module can perform its operation whenever it has a data available. 3) Since we are instantiating different modules for each layer, we can customize the parallelization factors of each module based on the available workload of their respective GCN layer. 4) As the adjacency matrix of a graph does not change across different layers, we can read the edges from the global memory only once for the first layer and reuse them for the subsequent ones by transferring them through the on-chip FIFOs.

\subsection{Extending the Support for Sparse Computation} \label{sec:gcn-sparse}
The input node embedding to the first layer of GCN usually contains many zero elements since it often uses one-hot encoding for assigning initial vectors to nodes.
Furthermore, as shown in Fig.~\ref{fig:gcn}, at the end of each GCN layer, there is a ReLU unit. As a result, the matrix generated by each layer, which is the input to the next layer, is sparse. In fact, we saw $52\%$ and $47\%$ sparsity on average for the input to the second and the third layers of GCN in SimGNN for randomly drawn graphs from our target dataset. 
As a result, the Feature Transformation step also needs to have the support for sparse computation.
To reduce the number of operations, we prune the zero elements and only pass the non-zero ones to the next layer. As the updates to the output buffer may come in random cycles, it is necessary to store the buffer containing the partial results on-chip to enable random access. For the same reason, we pack the node features with their address which includes their row and column ID. Packing the elements with their address helps to make the dispatch unit simpler since each SIMD PE is free to work with any data and knows which partial result should be updated; hence, there is no need to take special considerations to navigate the data to the correct PE. We only need to make sure that at all the times each SIMD PE is working with a different memory bank. We employ an arbiter for this matter, as explained below.

\begin{figure}[H]
	\centering
	\includegraphics[width=0.7\columnwidth]{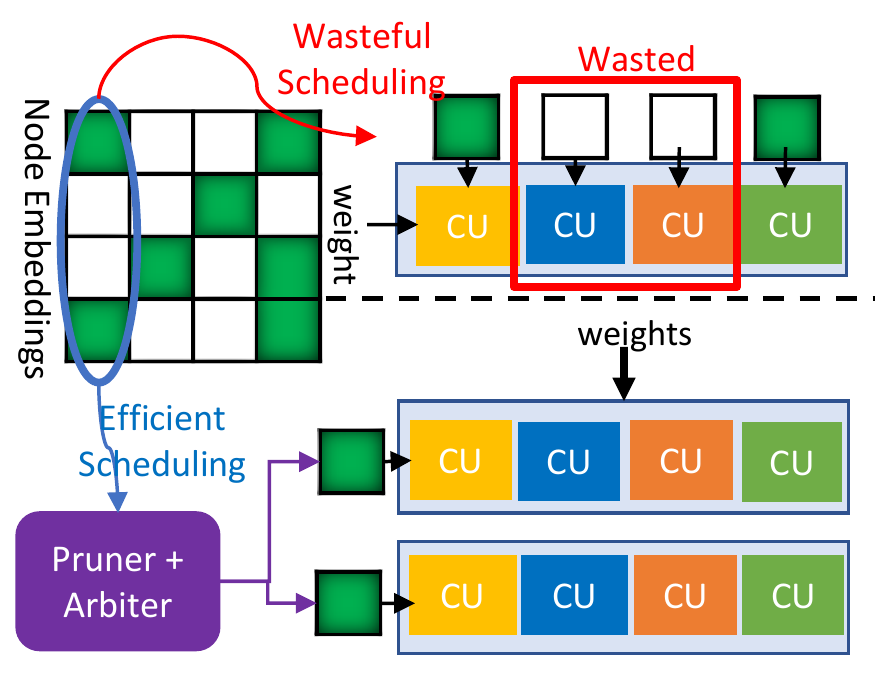} 
	\caption{The benefit of streaming the node embeddings and mapping the weights to the SIMD dimension for supporting sparse computation.}
	\label{fig:sched-pref}
\end{figure}

As mentioned in Section~\ref{sec:gcn_baseline}, to reduce the number of RAW dependencies, we chose to stream the node embedding matrix and broadcast it to different computation units (CU) which read the weight matrix as a batch. Since the node embedding is a sparse matrix, reading it as a stream facilitates the pruning mechanism we employ and enables us to distribute the workload more efficiently. Fig.~\ref{fig:sched-pref} demonstrates a toy example illustrating the benefit of this scheduling. The colored squares show the non-zero elements of the node embedding matrix. By mapping the weights, which are non-zero, to the SIMD dimension, all the CUs in the PE would execute useful operations and we can skip all the operations involving a zero node embedding.

When skipping the zero node embeddings, the dependency distance for output elements may change dynamically since the number of non-zero inputs between the updates to the same location can be different. Even though the scheduling discussed in Section~\ref{sec:gcn_baseline} increases the dependency distance as much as possible by doing all the MAC operations when a non-zero input is encountered (each non-zero element would fill $\frac{f_{out}}{SIMD_{FT}}$ cycles of the dependency window), there still may be some cases where the dependency distance is less than \textit{L} (the latency of our function unit (FU) causing the dependency) after this optimization. Instead of setting the II to \textit{L} to ensure correctness, we first insert \textit{L} registers to store the partial results of FU at the end of each of its pipeline stages; hence, we can schedule a new set of operations at each clock cycle (II=1). 
There may be cases where the new scheduled operations want to update a location whose old value is still in the registers and have not updated the buffer. To ensure the correctness, we add a control unit which keeps track of the last cycle that each of the output locations was updated. If the number of cycles between two updates to the same location is less than \textit{L}, the control unit will insert bubbles into the pipeline until the previous update is committed. 

\begin{figure}[!htb]
	\centering
	\includegraphics[width=\columnwidth]{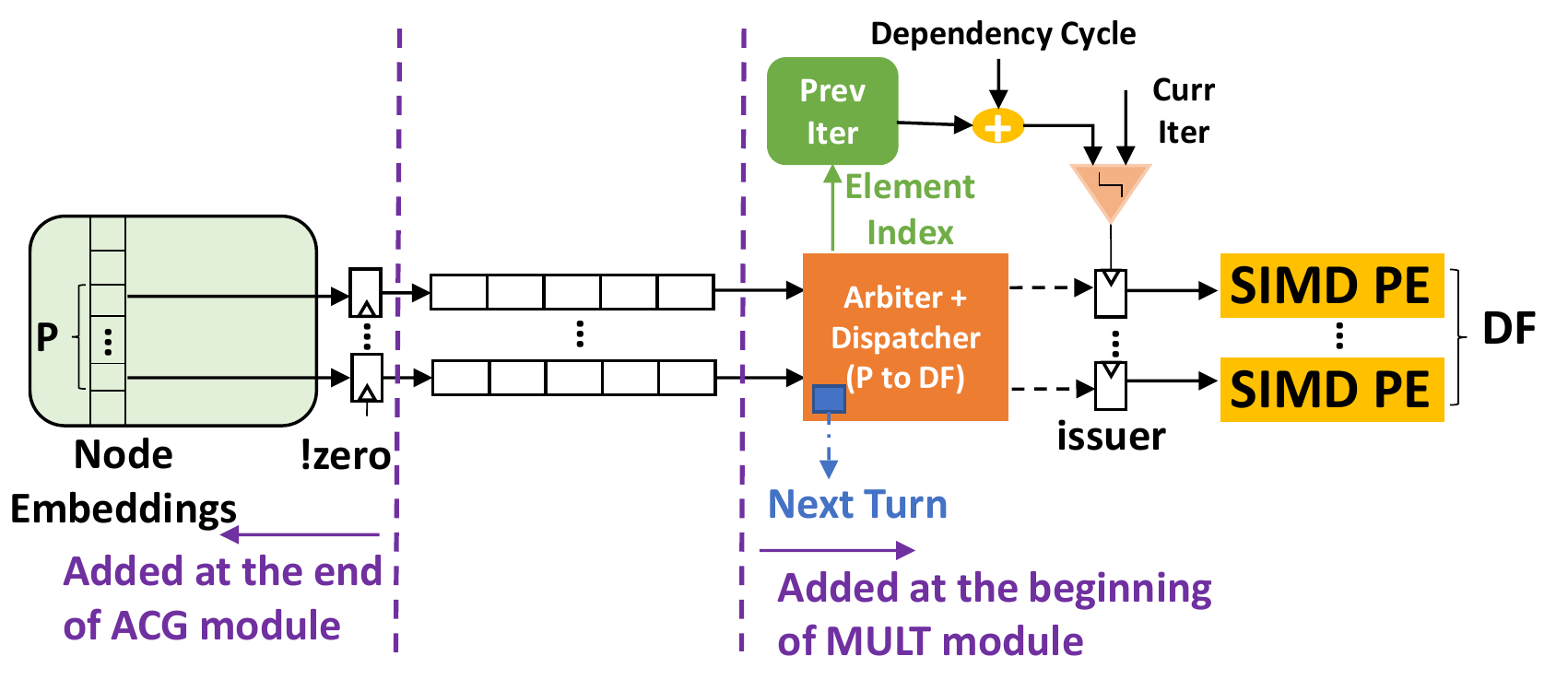} 
	\caption{Architecture support for sparse computation in Feature Transformation step.}
	\label{fig:sparse-comp}
\end{figure}

We insert a unit for pruning zeros at the end of the Aggregation step in \textit{ACG} module. As Fig.~\ref{fig:sparse-comp} demonstrates, at each cycle, we evaluate \textit{P} elements of the node embeddings and pass each to a FIFO if it is not zero. To prevent deadlocks in the system, we use non-blocking read or writes for these FIFOs and only write (read) elements when the FIFO is not full (empty). The \textit{MULT} module of the Feature Transformation step of the next layer takes the \textit{P} FIFOs as the input and uses an arbiter to fetch, at most, \textit{DF} of them ($DF <= P$) for passing to \textit{DF} SIMD PEs. An arbiter keeps track of the FIFO whose turn it is to be read first in the next cycle. It then uses a round-robin ordering for dispatching the elements from the non-empty FIFOs. After dispatching the inputs, it checks for the RAW dependency by scanning the \textit{prev iter} buffer which contains the last cycle when each element was seen as the input. If the distance was less than \textit{L}, it will insert bubbles in the pipeline until the previous input has committed its update to avoid the conflict on II. 
If there is no dependency, for each memory bank no more than one element from the dispatched inputs will be issued to a SIMD PE and the current cycle number will be stored in \textit{prev iter} buffer for that input. 

\begin{table}[!tbh]
\footnotesize
\centering
\caption{Summary of architecture parameters for the accelerator of each GCN layer in \mbox{\framework}.}
\label{tbl:design-parameters}
\begin{tabular}{|c|c|}
\hline
\textbf{Design Parameter} & \textbf{Explanation}  \\ \hline\hline
$SIMD_{FT}$ & SIMD factor of the FT step\\
\hline
$SIMD_{Agg}$ & SIMD factor of the Aggregation step\\
\hline
DF & Duplication factor of the PEs in FT step\\
\hline
P & Number of input FIFOs to the arbiter\\
\hline
\end{tabular}

FT: Feature Transformation
\end{table}

To recap, the \mbox{\framework} architecture provides a flexibility in choosing the parallelization rates. Table~\ref{tbl:design-parameters} lists the parameters that can be tuned for each GCN layer based on its workload. The SIMD factors correspond to feature-level parallelization, while $DF$ and $P$ map to node dimension. 

\section{\mbox{\framework} Application to Graph Matching} \label{sec:method}
In Section~\ref{sec:arc_gcn}, we proposed an architecture for GCN specialized for small graphs. In this section, we extend our architecture to accelerate an end-to-end application, SimGNN. An end-to-end application introduces new computation patterns beyond GCN which requires customized processing units. We introduce a new level of parallelization in Section~\ref{sec:exp-batch}.

\subsection{SimGNN} \label{sec:simgnn}

\begin{figure}[!phtb]
	\centering
	\includegraphics[width=\columnwidth]{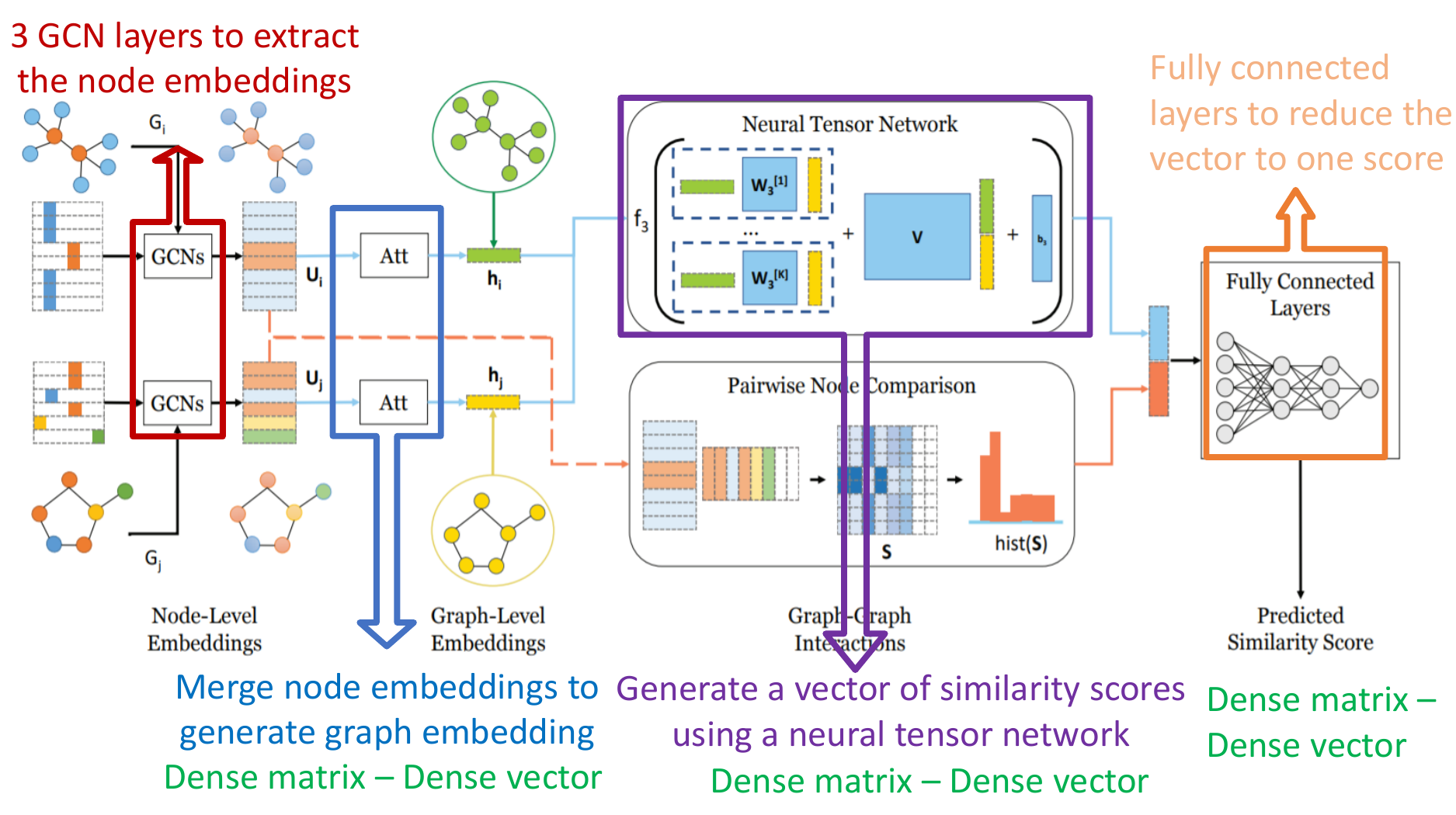} 
	\caption{The SimGNN pipeline as depicted in~\cite{bai2019simgnn}.}
	\label{fig:simgnn-pipeline}
\end{figure}

Bai et al. \cite{bai2019simgnn} proposed a neural-network-based approach to assign a similarity score to two graphs. Its computation pipeline consists of four major stages as depicted in Fig.~\ref{fig:simgnn-pipeline}. The first stage is made up of three layers of GCN to extract the \textit{node embeddings} $H \in \mathbb{R}^{|V|\times F}$ where $F$ is the number of features of the last layer. In the second stage, it uses a \textit{Global Context-Aware Attention layer (Att)} to combine the node embeddings and generate a single embedding per graph $h_G \in \mathbb{R}^{F}$. 
For this matter, at first, the Att stage computes the global context by taking an average of the node embeddings followed by a non-linear transformation: $c = tanh(\frac{1}{|V|}W_{Att}\Sigma_{n = 1}^{|V|} h_n)$ where $W_{Att} \in \mathbb{R}^{F\times F}$ is a learnable \textit{weight matrix}. Then, it calculates the importance of each node by constructing an \textit{attention weight} $a_n$ for them to denote their similarity score to the global context by 
computing: $a_n = \frac{1}{1 + exp(h_n^T \cdot c)}$. Finally, the graph embedding can be calculated by taking a weighted sum of the node embeddings using the attention weights. The computation in this stage can be summarized as follows:
\begin{align} \label{eq:att}
    \begin{split}
        h_G = \sum_{n = 1}^{|V|} \sigma(h_n^T \cdot tanh(\frac{1}{|V|}W_{Att}\Sigma_{n = 1}^{|V|} h_n)) \cdot h_n
    \end{split}
\end{align}
where $\sigma(\cdot)$ denotes the sigmoid function to produce $a_n$. The time complexity of this stage can be seen to be $O(|V|F)$.

The third stage is a \textit{Neural Tensor Network (NTN)} that calculates a vector of similarity score between the two graphs:
\begin{align} \label{eq:ntn}
    \begin{split}
        s(h_{G_1}, h_{G_2}) = \sigma(h_{G_1}^{T}W_{NTN}^{[1:K]}h_{G_2} + V \cdot concat(h_{G_1}, h_{G_2}) + b)
    \end{split}
\end{align}
where $W_{NTN}^{[1:K]} \in \mathbb{R}^{F\times F \times K}$, $V \in \mathbb{R}^{K \times 2F}$, and $b \in \mathbb{R}^{K}$ are learnable \textit{weight tensor}, \textit{weight matrix}, and \textit{bias vector}, respectively. $K$ is a hyper-parameter that controls the number of similarity scores. The time complexity of this stage is $O(F^2 K)$ where F denotes the dimension of the graph level embedding. In the last stage, a standard fully connected network is used to gradually reduce the similarity vector to only one score. 

Since the core computation stage of SimGNN consists of GCN layers, the challenges mentioned in previous sections still exist here. Furthermore, the rest of the stages make use of exponential and hyperbolic tangent functions which are expensive to have on FPGA that can limit the rate of parallelism for them. By comparing the computation complexity of these stages, with GCN (Section~\ref{sec:gcn}), we can see that the GCN step is the most computation-intensive step; hence, when pipelining all the stages together, the accelerator will be bottlenecked by the GCN step. Therefore, we do not aggressively parallelize the rest of the steps. We rather focus on reducing their resource utilization.


\subsection{Att Architecture} \label{sec:arc_att}
The SimGNN pipeline applies the GCN stage to two graphs for each comparison query. Instead of duplicating the architecture in Fig.~\ref{fig:high-level-gcn} twice, we process the graphs serially and reuse the GCN module for the two input graphs in the query. Reusing the GCN module enables us to map the design to smaller FPGAs as well. Note that the GCN stage is the most resource-hungry one of them all. 
We improve the performance for processing one query by overlapping the \textit{GCN} computation of one graph with the \textit{Att} computation of the other one. Thus, the total performance will be bottlenecked with the performance of GCN. As a result, we focus on reducing the area and reusing the resources for \textit{Att}. Let $H \in \mathbb{R}^{F \times |V|}$ be the transposed result of the GCN stage of SimGNN, then, its n-th column ($x_n$) will be the vector $h_n$ in Eq.~\ref{eq:att}. In computing $v = W_{Att}\Sigma_{n=1}^{|V|} h_n$, we first have to add $h_n$ vectors and then do a matrix-vector multiplication (MVM). 
Instead of instantiating separate adders for the first additions and the ones in MVM, we rewrite the equation as follows to reuse the adders:
\begin{align}
    \begin{split}
        v = W_{Att}\Sigma_{n=1}^{|V|} h_n = \Sigma_{n=1}^{|V|} W_{Att} h_n = sum(W_{Att} \cdot H, 2)
    \end{split}
\end{align}
where $sum(W_{Att}.X^T, 2)$ denotes the reduction of the result matrix across its second dimension (columns), meaning that all the multiplications associated with a column of $H$ should be added together.

\begin{figure}[!htb]
	\centering
	\includegraphics[width=0.83\columnwidth]{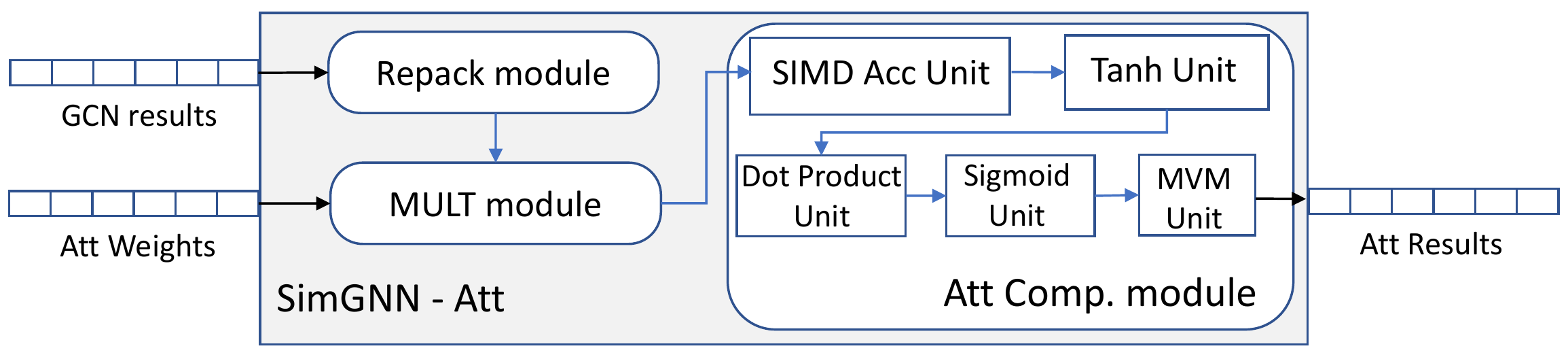} 
	\caption{Architecture overview of the second stage of SimGNN in \mbox{\framework}: Att}
	\label{fig:arc-att}
\end{figure}

Fig.~\ref{fig:arc-att} demonstrates an overview of the \textit{Att} module. As in the GCN stage, we divide the matrix multiplication to two different modules, one responsible for multiplications and the other for additions. Again, we use SIMD PEs to implement these modules. However, the SIMD factor here can be set to a different value compared to the GCN stage since they have different computation complexities. The \textit{Repack} module is responsible for adjusting the output of GCN with the SIMD factor of this stage. We use the implementation of \textit{tanh} and \textit{exp} functions available in \textit{Xilinx HLS Math} library since they are already optimized. Note that the last summation in Eq.~\ref{eq:att} can be seen as $H\times a$ where $a \in \mathbb{R}^{|V|}$ contains the sigmoid results. Hence, we use a matrix vector multiply (MVM) unit at the end.

\vspace{-0.1in}
\subsection{NTN + Fully-connected Network Architecture} \label{sec:arc_ntn}
\begin{figure}[!htb]
	\centering
	\includegraphics[width=0.88\columnwidth]{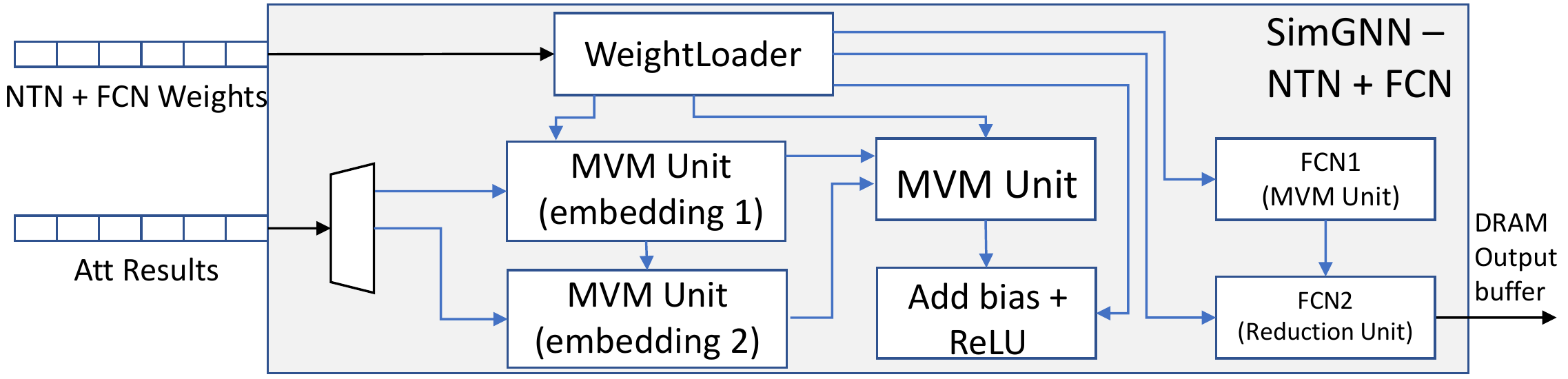} 
	\caption{Architecture overview of the last two stages of SimGNN in \mbox{\framework}: NTN and FCN.}
	\label{fig:arc-ntn}
\end{figure}

The computation in the NTN stage is rather simple since it is a series of fixed-size MVMs followed by a bias addition and an activation function. Furthermore, the layers of the fully-connected network (FCN) in the last stage either need an MVM unit or a reduction tree to lower a vector to a scalar. Like the previous stages, we implement all the sub-modules of these two stages in a dataflow-manner. Fig.~\ref{fig:arc-ntn} depicts the architecture of these two steps.

\subsection{Putting It All Together} \label{sec:arc_simgnn}
Finally, we develop the SimGNN architecture by connecting the modules described in the previous sections. The whole computation pipeline is implemented as a three-level dataflow architecture. The first two levels resemble an inter-stage pipelining while the last one is for intra-stage pipelining. The first level enables a task-level parallelization by grouping the graph-related steps, the \textit{GCN} (Section~\ref{sec:arc_gcn}) and \textit{Att} (Section~\ref{sec:arc_att}) modules, and overlapping them with the rest, \textit{NTN\_FCN module} (Section~\ref{sec:arc_ntn}). The second level of dataflow architecture overlaps the \textit{GCN} stage with the \textit{Att}. Finally, the last level applies dataflow architecture to each of the the \textit{GCN}, \textit{Att}, and \textit{NTN\_FCN} modules as shown in Fig.~\ref{fig:high-level-gcn},~\ref{fig:arc-att}, and~\ref{fig:arc-ntn}, respectively. 
The \textit{Pre-fetcher} of the \textit{GCN} module reads the weights and bias of the rest of the stages as well and distributes them to the right place.

We apply three optimizations for reducing the off-chip communication latency: 1) each input buffer can be mapped to a different DRAM bank or HBM channel to enable parallel access to them, 2) the available global memory bandwidth is fully utilized by applying memory coalescing. Memory burst is also applied to amortize the initialization overhead, 3) the modules that access the global memory are overlapped by the computation modules by implementing the accelerator as a dataflow architecture. 

\section{Experimental Results} \label{sec:eval}
In this section, we first review our target dataset in Section~\ref{sec:exp-benchmark} and present the experimental setup in Section~\ref{sec:exp_setup}.
We then evaluate the effects of our optimizations to the GCN architecture (Section~\ref{sec:arc_gcn}) step-by-step in Section~\ref{sec:exp}.
The whole pipeline of \mbox{\framework} is then evaluated in Section~\ref{sec:exp-overall} on three different FPGAs that differ in the amount of resources available and the type of global memory they employ.
Finally, we compare the performance of \mbox{\framework} to CPU and GPU implementation and demonstrate the superiority of our design.
Related work on general GCN acceleration, sparse CNN, and sparse matrix multiplication will be discussed in Section~\ref{sec:rel-work}.

\subsection{Benchmark}
\label{sec:exp-benchmark}
We consider a publicly available dataset, AIDS~\cite{aids}, for benchmarking our design.
AIDS is a real-life graph dataset containing 42,687 antivirus chemical compounds gathered by the Developmental Therapeutics Program at NCI/NIH. 
The graphs in AIDS have 25.6 nodes and 27.6 edges on average. 
We randomly select 10,000 pairs to form 10,000 queries.
The kernel time and end-to-end (E2E) time reported in this section are the average of all queries.

\subsection{Experimental Setup} \label{sec:exp_setup}
The \mbox{\framework} architecture is described using Vivado HLS C++~\cite{vivado}. The design is synthesized and implemented using Xilinx Vitis 2019.2~\cite{vitis} on three different target platforms: Xilinx Alveo U50, Xilinx Alveo U280, and Xilinx Kintex UltraScale+ KU15P. The first two are equipped with HBM2~\cite{hbm} and, ideally, can achieve a bandwidth of 316 GB/s (460 GB/s) with a TDP of 75W (225W)~\cite{u50, u280}; while the last one utilizes DDR4 as the global memory. Table~\ref{tbl:resource_fpgas} compares the hardware resources of these boards. 
For comparison to CPU and GPU, the PyTorch-based implementation of SimGNN from~\cite{simgnn-github} is used that is built using the state-of-the art PyTorch Geometric (PyG) library~\cite{pyg} which is commonly used as a baseline by previous works~\cite{yan2020hygcn, geng2020awb, liang2020engn}. For the Aggregation step, PyG exploits sparsity and edge-level parallelism by adapting the PyTorch Scatter library~\cite{scatter}. For the Feature Transformation step, it uses Intel MKL~\cite{mkl} and NVIDIA cuBLAS library~\cite{cublas} for CPU and GPU respectively, making it a reasonable and optimized baseline.
The target CPU in our experiments is Intel(R) Xeon(R) CPU E5-2699 v4 running at 2.2 GHz.
For testing on GPU, we use an AWS p3.2xlarge instance which has an NVIDIA V100 GPU.

\begin{table}[!tbh]
\footnotesize
\centering
\caption{Properties of the FPGAs used in this paper.}
\label{tbl:resource_fpgas}
\begin{tabular}{|c|c|c|c|c|c|c|}
\hline
Platform & \makecell{BRAM \\ (Mb)} & \makecell{LUT \\ (K)} & \makecell{FF \\ (K)}  & DSP  & \makecell{URAM \\ (Mb)} & \makecell{Max BW \\ (GB/s)}   \\ \hline\hline
KU15P & 34.6 & 523 & 1045 & 1968 & 36 & 19.2 \\
\hline
U50 & 47.3 & 872 & 1743 & 5952 & 180 & 316 \\
\hline
U280 & 70.9 & 1304 & 2607 & 9024 & 270 & 420 \\
\hline
\end{tabular}
\end{table}

\subsection{Impact of GCN Architecture Optimizations} \label{sec:exp}

\subsubsection{Inter-Layer Pipelining:}

\begin{table*}[!tbh]
  \footnotesize
  \centering
  \caption{Impact of GCN architecture optimizations on U280. The meaning of design parameters is summarized in Table~\ref{tbl:design-parameters}. Baseline shows a single set of design parameters because it uses the same hardware for all layers.}
  \label{tbl:hardware-opt}
  \resizebox{\linewidth}{!}{
    \begin{tabular}{|c|c|c|c|c|c|c|c|c|}
      \hline
      \multirow{2}{*}{Architecture} & \multicolumn{4}{c|}{Design Parameters (L1 / L2 / L3)} & LUT / FF / DSP / & Freq. & \multirow{2}{*}{Kernel (ms)} & \multirow{2}{*}{Kernel $\times$ DSP}                                                      \\
      \cline{2-5}
                                    & $SIMD_{FT}$                                           & $SIMD_{Agg}$     & DF    & P                            & BRAM / URAM (\%)                     & (MHz) &                      &                     \\
      \hline\hline
      Baseline                      & 16                                                    & 32               & 8     & ---                          & 9.8  / 7.7 / 7.4 / 6.8 / 0           & 265   & 0.599 ($1\times$)    & 4.46 ($1\times$)    \\
      \hline
      +Inter-Layer Pipeline         & 32/16/16                                              & 32/32/16         & 8/8/8 & ---                          & 14  / 12   / 18  / 3.6 / 2.5         & 271   & 0.383 ($1.56\times$) & 6.74 ($0.66\times$) \\
      \hline
      +Extended Sparsity            & 32/32/16                                              & 32/32/16         & 2/1/1 & 8/2/2                        & 4.8 / 6.0  / 4.4 / 4.8 / 3.1         & 300   & 0.264 ($2.27\times$) & 1.15 ($3.88\times$) \\
      \hline
    \end{tabular}
  }
\end{table*}

Table~\ref{tbl:hardware-opt} shows the resource usage and performance of the \mbox{\framework} architecture when accelerating three GCN layers of SimGNN on the U280 FPGA.
The baseline uses the same hardware for all GCN layers.
With inter-layer pipeline added, all 3 GCN layers run in parallel as a coarse-grained pipeline.
Since each layer utilizes different pieces of hardware,
we can customize the design parameters to match the throughput of each layer.
As a result, the 3 layers require 2.4$\times$ more DSPs compared with the baseline.
Although BRAM usage appears to be smaller,
more URAM is used with inter-layer pipelining since we distribute the required resources to obtain to a better frequency. The total storage usage is increased in order to store all the buffers for different GCN layers and the FIFOs for the intermediate results between them.
The GCN kernel time is reduced by 36\% with inter-layer pipelining added to the baseline.
However, if we look at the latency-area product metric, i.e., Kernel$\times$DSP, we can see that the performance improvement does not catch up with the computation units (DSP) increment, suggesting potential for further optimizations.

\subsubsection{Extending Sparsity to Feature Transformation:}

Indeed, we can extend sparsity support to the Feature Transformation step in GCN to improve both the kernel time and the latency-area product metric, as explained in Section~\ref{sec:gcn-sparse}. Although using \textit{P} queue helps the arbiter fetch non-zero elements more frequently, it may still not be enough to dispatch data to all the \textit{DF} PEs. Furthermore, by increasing the \textit{DF}, we may need to insert more bubbles in the pipeline to avoid RAW dependency since it reduces the number of cycles between the updates to the same location. As a result, there is a trade-off in choosing the right \textit{DF} for each layer. Since it is a highly workload-dependent decision, we employ profiling results for setting each of the parallelization factors. 

The best parallelization factors are summarized in Table~\ref{tbl:hardware-opt}.  
When DF is set to 1, we no longer need to have separate banks in the row dimension of the buffers which can lessen the number of needed memory blocks. 
This makes it more efficient to use dense memory blocks (BRAM and URAM) as opposed to LUTs for the buffers, thus BRAM and URAM usage are increased.
As Table~\ref{tbl:hardware-opt} shows, extending sparsity to feature transformation over inter-layer pipeline has further reduced the kernel time by 31\%, effectively achieving a 2.27$\times$ speedup over the baseline, while decreasing the DSP usage by 4.09$\times$. The results clearly suggest that, since this is a memory-bounded application, throwing more resources to the architecture is not helpful. Instead, the memory access latency should be reduced and the computation units shall be used more efficiently.  
Since a large number of zero elements and required DSPs are excluded, the latency-area metric Kernel$\times$DSP is greatly improved by 3.88$\times$ over the baseline.

\subsection{End-to-end Acceleration of SimGNN} \label{sec:exp-overall}
\subsubsection{Flexibility of Mapping to Different FPGAs:}
\begin{figure}[!htb]
	\centering
	\includegraphics[width=\columnwidth]{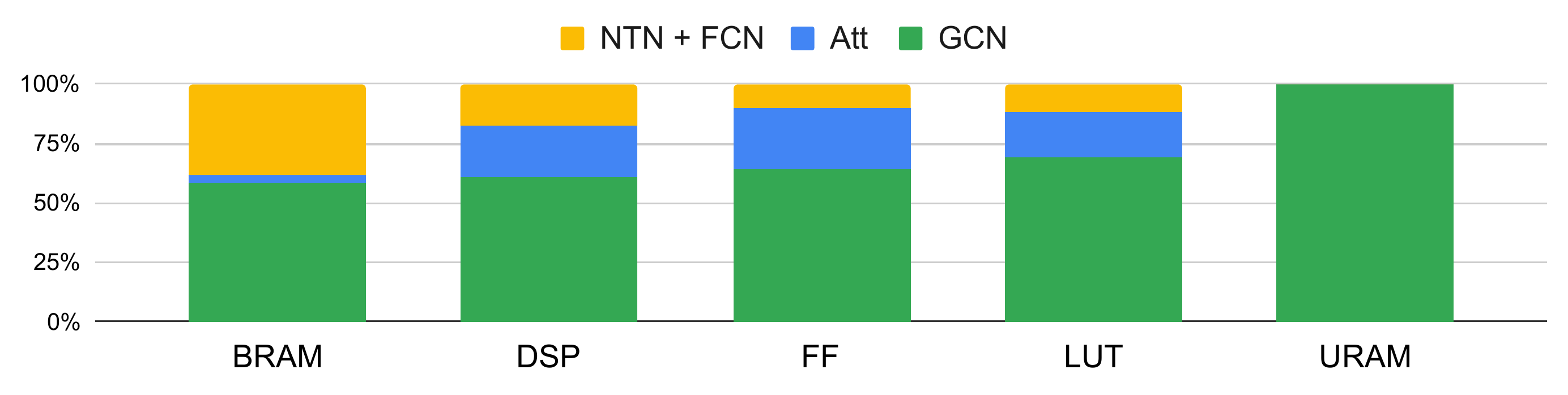} 
	\caption{Resource breakdown of the whole pipeline of SimGNN on U280.}
	\label{fig:res-breakdown}
\end{figure}

We implement the whole pipeline of SimGNN on 2 HBM FPGAs and KU15P that uses DDR memory. Fig.~\ref{fig:res-breakdown} compares the resource breakdown of the modules at the top hierarchy of our design when mapped to U280. As the figure exhibits, we allocate most of the resources to the GCN stage as it is the computation-intensive part of the network.
Table~\ref{tbl:fpgas} shows the resource usage and performance for the three FPGA platforms.
We can see that the kernel runs faster on HBM FPGAs compared to KU15P. This is mainly due to the fact that HBM FPGAs can achieve a better frequency as they have more resources and the Vitis tool has more freedom in PnR to optimize the timing. The fact that the multiplication and addition units have different latencies on these boards (5 and 8 cycles for KU15P; 4 and 7 cycles for U280, respectively) further increases the difference in the runtimes. In fact, the cycle count of the same kernel when it uses different types and number of banks for global memory is almost the same. This suggests that after our optimizations the bottleneck is no longer at the memory level.
Besides, we can see that the end-to-end time is much larger than the kernel time (runtimes measured on the host using CPU clock). This is mainly because of the overhead of launching the kernel and PCIe transfers since all the pre-processing steps take less than 5\% of the end-to-end time. 
This suggests the potential for batching queries to further improve the end-to-end throughput (Section~\ref{sec:exp-batch}). 

\begin{table}[!tbh]
  \footnotesize
  \centering
  \setlength\tabcolsep{2.65pt}
  \caption{Performance and resource utilization of a \mbox{\framework} design on different target FPGAs.}
  \label{tbl:fpgas}
  \begin{tabular}{|c|c|c|c|c|c|}
    \hline
    FPGA  & \makecell{ LUT / FF / DSP /                              \\ BRAM / URAM (\%)}  & \makecell{Freq. \\ (MHz)} & \makecell{Kernel \\ (ms)}  & \makecell{E2E\\(ms)} & \makecell{E2E\\(query/s)}  \\ \hline\hline
    KU15P & 34 / 29 / 35 / 30 / 23    & 201 & 0.786 & 1.135 & 881  \\
    \hline
    U50   & 17 / 16 / 12 / 16 / 4.7   & 279 & 0.423 & 0.538 & 1858 \\
    \hline
    U280  & 11 / 10 / 7.7 / 10 / 3.1  & 290 & 0.327 & 0.509 & 1965 \\
    \hline
  \end{tabular}
  E2E: End-to-End
\end{table} 



\begin{table}[!tbh]
\footnotesize
\centering
\caption{Performance comparison of running SimGNN on different hardwares. The CPU and GPU runtimes are based on state-of-the-art PyG~\cite{pyg} implementation. Low computation complexity of small graphs and the coarse-grained execution of GPUs impact the efficiency of GPU significantly.}
\label{tbl:fpga-cpu-gpu}
\begin{tabular}{|c|c|c|c|c|c|}
\hline
Platform & \makecell{Max BW \\ (GB/s)} & \makecell{Kernel \\ (ms)} & \makecell{E2E \\ (ms)} & \makecell{Speedup \\ (Over CPU)}  & \makecell{Speedup \\ (Over GPU)}    \\ \hline\hline
KU15P & 19.2 & 0.786 & 1.135 & 8.2 &  12.1 \\
\hline
U50 & 316 & 0.423 & 0.538 & 17.2 &  25.5 \\
\hline
U280 & 460 & \textbf{0.327} & \textbf{0.509} & \textbf{18.2} &  \textbf{26.9 }\\
\hline
PyG-CPU & 76.8 & 5.85 & 9.27 & 1 &  1.5 \\
\hline
PyG-GPU (V100) & 900 & 9.68 &  13.7  &  0.68  & 1\\
\hline
\end{tabular}

BW: Bandwidth, E2E: End-to-End
\end{table}

\subsubsection{\mbox{\framework} vs CPU and GPU:} \label{sec:comparisons}
We test the performance of the whole pipeline of SimGNN on the CPU and GPU described in Section~\ref{sec:exp_setup}. In this section, we are assuming that the inputs are already stored in the host memory, and we want to offload the graph comparison queries to either of the target platforms. The goal is to compare the performance of these platforms for processing a graph matching query. Table~\ref{tbl:fpga-cpu-gpu} summarizes the results. The results are averaged over 10,000 queries on each of the platforms. 
The queries are started sequentially, and the end-to-end time of all the platforms is the time interval between two consecutive queries are started. This contains the runtime for any pre-processing steps as well.
For FPGA and GPU, it also includes the host-kernel communication via the PCIe link, writing data to FPGA/GPU's global memory, kernel computation, reading the results from that, and the overheads for using the APIs (OpenCL for FPGA and PyTorch for CPU/GPU). We use the end-to-end time for comparison since these overheads are inevitable and should be accounted for. 
The kernel time on CPU/GPU is measured with the PyTorch profiler.

The results demonstrate that our FPGA solution can outperform both CPU and GPU significantly. As discussed in Section~\ref{sec:intro}, this is partly because of the dynamic load balance and the irregular memory access of the graph structure. Furthermore, since we target small graphs, it results in extreme under-utilization of GPU.
In fact, the profiling results indicate that the GPU utilization does not go higher than $6\%$ and, for the most part, the PyG-GPU only uses 1 streaming multiprocessor (SM) since the matrices are small. Because of this and the fact that GPU runs at a lower frequency (1.3GHz) compared to CPU (2.2 GHz), the GPU version of this application is even slower than the CPU. The \textit{nvprof} profiling results show that PyG-GPU runs 225 kernels for accelerating this application that on average have 4.6 $KFLOPs$. With this low computation intensity, the overhead of running the kernel (such as \textit{cudaLaunchKernel}) is larger than the actual kernel runtime that greatly impacts the GPU performance. 
Designing the GPU kernel manually can alleviate some of these shortcomings, but the underlying problem still exists due to the coarse-grained execution model of GPU. In contrast, our FPGA solution suffers from the kernel initialization overheads only once since we develop a deep pipeline across all stages of the computation by fusing them in one kernel. This pipelining has several other benefits as explained in Section~\ref{sec:gcn-TLP}.
\begin{figure}[!htb]
	\centering
	\includegraphics[width=\columnwidth]{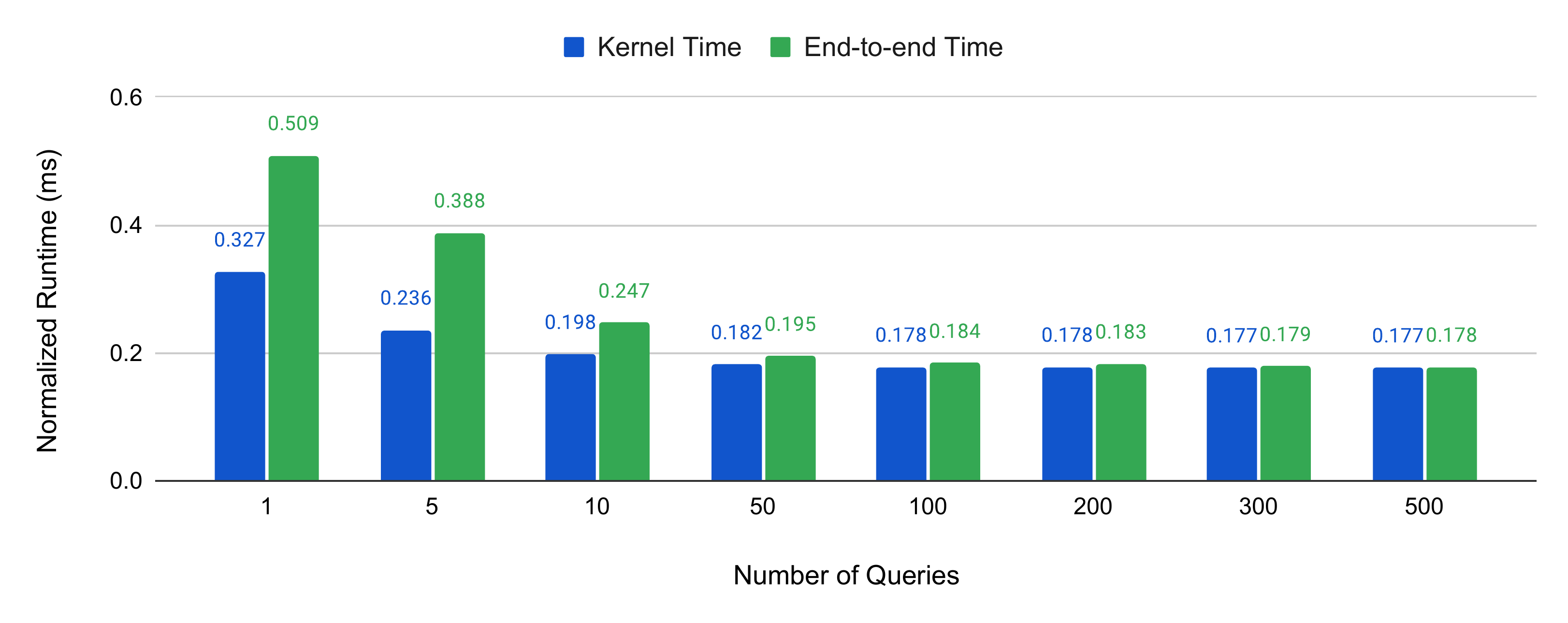} 
	\caption{Effect of batching queries. Runtime numbers are the average over all queries.}
	\label{fig:queries}
\end{figure}

\subsubsection{Batching Queries on FPGA:}
\label{sec:exp-batch}
The difference in the end-to-end time and the kernel time on FPGA is mainly due to both the DMA transfers and overhead of OpenCL APIs. In fact, our experiments using the Profile Summary provided by Vitis Runtime library~\cite{vitis_manual_ug1393} show that OpenCL APIs can take 10-100 $\mu$s which is comparable to the kernel execution for one query.
Therefore, \mbox{\framework} implements batching graph queries to amortize these overheads.
We can send the data for processing multiple queries to FPGA, process each query serially, and send the results back together.
Fig.~\ref{fig:queries} demonstrates the results of this experiment on U280 board.
The results suggest that by transferring the data for \textasciitilde 300 queries, we can amortize the setup time and achieve a $2.8\times$ speedup.

Batching queries makes it possible to further improve the throughput by adding another level of parallelization.
When targeting the HBM FPGAs, we use 4 of the HBM PCs for processing one query of graph matching. As each HBM FPGA has 32 PCs that can be accessed independently, we can further improve the performance by replicating the logic of \mbox{\framework} for one query to process up to 8 different queries in parallel as long as we have enough resources available. 
Table~\ref{tbl:resource_fpgas} illustrates that the available resources allow us to instantiate 6 \mbox{\framework} pipelines with U280 before hitting the 80\% resource usage upper-bound.
This does not change the latency of each graph query, but it would further increase the throughput by $6\times$ resulting in a throughput of 33522 query/s on U280.

\section{Related Work} \label{sec:rel-work}

\textbf{GCN Accelerators:} Because of the popularity of GCN, there is a growing interest in developing an accelerator for it~\cite{zeng2020graphact, yan2020hygcn, geng2020awb, zhang2020accelerating, liang2020engn}. As summarized in Table~\ref{tbl:comparison}, HyGCN~\cite{yan2020hygcn}, GraphACT~\cite{zeng2020graphact}, and Prasanna et al.~\cite{zhang2020accelerating} develop a fixed hardware for all the layers of GCN and process them sequentially. This is an undesirable feature particularly when we target small graphs. 
In fact, our baseline architecture, in which we reuse the same architecture for all the GCN layers, exploit only the sparsity of the Aggregation step, treat the Feature Transformation step as a regular matrix multiplication, and employ a 2D computation unit for it, has the same design principles as these works. However, as the experimental results in Table~\ref{tbl:hardware-opt} show, not only should we execute the GCN layers in a pipelined fashion, but also we should exploit the sparsity of the node embeddings to enhance both area and performance.
In addition, GraphACT and Prasanna et al.~\cite{zhang2020accelerating} mainly rely on \textit{redundancy reduction} in the Aggregation step to decrease the number of operations by pre-computing the repeated aggregations. This is possible only when the adjacency matrix is \textit{binary}. However, not all GCNs can benefit from this feature since they typically work with the \textit{normalized} adjacency matrix meaning that they need \textit{weighted} additions in this step. AWB-GCN~\cite{geng2020awb} proposes an architecture that supports inter-layer pipelining and considerations for sparsity of the node embeddings for accelerating GCN. However, partitioning the computation by the nodes in their approach complicates the design of the task distributor since the node embeddings are sparse and special consideration is needed to prevent PEs from doing unnecessary operations on the zero elements. On the other hand, feature-level parallelization deals better with workload imbalance as we discussed in Section~\ref{sec:arc_gcn}. In addition, AWB-GCN is developed for large graphs and is based on the inner-product matrix multiplication, whereas, as we discuss in Section~\ref{sec:gcn_baseline}, the outer-product multiplication is preferred to reduce the RAW dependency especially when we are targeting small graphs and do not have enough (non-zero) nodes to fill the dependency window.

Moreover, when targeting small graphs, we need more optimizations. For example, it is important to reduce the number of accesses to the global memory as much as possible. This is not the focus of the previous works. Besides, in addition to fitting the design for the whole network into an FPGA, we have the option to process parallel queries of the network. Table~\ref{tbl:comparison} summarizes the differences of our design compared to the aforementioned works on GCN acceleration.

\textbf{SpMM and SCNN Accelerators:} Apart from the works focusing on GCN, there has been a lot of research on sparse matrix multiplication (MM) either for pruned CNNs or normal  MM~\cite{srivastava2020matraptor,han2016eie, kung2019packing, parashar2017scnn, zhang2016cambricon, ding2017circnn, isca21-dual-side-sparse-tensor-core, hpca20-tensaurus, micro19-sparse-tensor-core}. They all rely on the fact that the sparse matrix is known offline and they can pre-process it. For example, EIE~\cite{han2016eie} propose a sparse matrix vector multiplier for the fully-connected layers. It reorganizes the sparse matrix in compressed sparse column (CSC) format and pre-loads that into on-chip memory. As another example, Kung et al.~\cite{kung2019packing} pre-process the data by merging multiple sparse columns of the weight matrix into one and pruning all the weights except for the most-significant ones. resulting in some accuracy loss. These approaches are not feasible for GCN in which the sparse matrix (i.e. the node embeddings) is generated \textit{while running} the algorithm; whereas, we proposed a technique to prune the zeros \textit{on-the-fly}.
\section{Conclusion}
In this paper, we analyzed and examined the optimization opportunities when GCN is applied to small graphs. We presented an efficient architecture, \mbox{\framework}, and developed an accelerator for SimGNN based on that as an end-to-end application. SimGNN is a neural-network-based graph matching algorithm for calculating an approximation of the GED between two graphs. 
 The computation disparity existing in the network calls for a customized accelerator. Besides, since the GPU has coarse-grained execution, we cannot have improvement beyond the optimizations applied for each phase since different phases are executed separately. However, on the FPGA side, we can exploit a deep pipeline across the phases by enabling a dataflow architecture. Not only does it help us reduce the global memory transactions, we can also eliminate the overhead of running different kernels. Furthermore, we showed that since this is a memory-bounded application, instantiating many computation units (as in GPU) is not beneficial. 
The experimental results demonstrate that \mbox{\framework} can outperform CPU and GPU by {\speedupcpu}$\times$ and {\speedupgpu}$\times$, respectively, because of the use of a very deep pipeline with different levels of parallelization.


\begin{acks}
We would like to thank Marci Baun for editing the paper. A significant part of this work was conducted while the first author was interning at Samsung Semiconductor Inc. It is also supported by the CAPA award jointly funded by NSF (CCF-1723773) and Intel (36888881), the RTML award funded by NSF (CCF-1937599), and CDSC industrial partners\footnote{https://cdsc.ucla.edu/partners/}.
\end{acks}


\bibliographystyle{ACM-Reference-Format}
\bibliography{main}

\end{document}